%% file: nsface_for_review.tex
\documentclass[10pt,twocolumn,letterpaper]{article}

\usepackage{iccv}
\usepackage{times}
\usepackage{epsfig}
\usepackage{graphicx}
\usepackage{amsmath}
\usepackage{amssymb}
\graphicspath{{figs/}}

\usepackage[pagebackref=true,breaklinks=true,letterpaper=true,colorlinks,bookmarks=false]{hyperref}

\iccvfinalcopy % *** Uncomment this line for the final submission
\def\iccvPaperID{1388} % *** Enter the ICCV Paper ID here

\ificcvfinal\fi
\begin{document}

%%%%%%%%% TITLE
\title{$L_{2}$-constrained Softmax Loss for Discriminative Face Verification}

\author{\parbox{16cm}{\centering
    {\large Rajeev Ranjan, Carlos D.  Castillo and Rama Chellappa}\\
    {\normalsize
    Center for Automation Research, UMIACS, University of Maryland, College Park, MD 20742\\}
    {\small \{rranjan1,carlos,rama\}@umiacs.umd.edu\\}}
%    \thanks{rranjan1,swamiviv,carlos,rama@umiacs.umd.edu}% <-this % stops a space
}

\maketitle
%\thispagestyle{empty}

%%%%%%%%% ABSTRACT
\input{abstract}
\input{introduction}

\input{relatedWork}

\input{motivation}

\input{method}
\input{results}
\input{conclusion}

\section{ACKNOWLEDGMENTS}

This research is based upon work supported by the Office of the Director of National Intelligence (ODNI), Intelligence Advanced Research Projects
Activity (IARPA), via IARPA R\&D Contract No. 2014-14071600012. The views and conclusions contained herein are those of the authors and should
not be interpreted as necessarily representing the official policies or endorsements, either expressed or implied, of the ODNI, IARPA, or the U.S. Government. The U.S. Government is authorized to reproduce and distribute reprints for Governmental purposes notwithstanding any copyright annotation
thereon.

{\small
\bibliographystyle{ieee}
\bibliography{nsfbib}
}

\end{document}

%% file: abstract.tex
%!TEX root = nsface_for_review.tex

\begin{abstract}
   In recent years, the performance of face verification systems has significantly improved using deep convolutional neural networks (DCNNs).  A typical pipeline for face verification includes training a deep network for subject classification with softmax loss, using the penultimate layer output as the feature descriptor, and generating a cosine similarity score given a pair of face images. The softmax loss function does not optimize the features to have higher similarity score for positive pairs and lower similarity score for negative pairs, which leads to a performance gap. In this paper, we add an $L_{2}$-constraint to the feature descriptors which restricts them to lie on a hypersphere of a fixed radius.  This module can be easily implemented using existing deep learning frameworks. We show that integrating this simple step in the training pipeline significantly boosts the performance of face verification. Specifically, we achieve state-of-the-art results on the challenging IJB-A dataset, achieving True Accept Rate of 0.909 at False Accept Rate 0.0001 on the face verification protocol. Additionally, we achieve state-of-the-art performance on LFW dataset with an accuracy of 99.78\%, and competing performance on YTF dataset with accuracy of 96.08\%. 
   
%The feature extraction and the similarity computation steps are usually decoupled which results in a performance gap.

%In recent years, the performance of face verification systems has significantly improved using deep convolutional neural networks (DCNNs). A typical pipeline for face verification includes training a deep network for subject classification with softmax loss, using the penultimate layer output as the feature descriptor, and generating a cosine similarity score given a pair of face images. In this paper, we add an $L_{2}$-constraint to the feature descriptors which restricts them to lie on a hypersphere of a given radius. This module can be easily implemented using existing deep learning frameworks. We show that integrating this simple step in the training pipeline significantly boosts the performance of face verification. Specifically, we achieve state-of-the-art results on the challenging IJB-A dataset, achieving True Accept Rates of 0.863 and 0.910 at False Accept Rates 0.0001 and 0.001 respectively on the face verification protocol. 

\end{abstract}

%% file: introduction.tex
%!TEX root = nsface_for_review.tex

\section{Introduction}

% What is the problem here?
Face verification in unconstrained settings is a challenging problem. Despite the excellent performance of recent face verification systems on curated datasets like Labeled Faces in the Wild (LFW)~\cite{huang2007labeled}, it is still difficult to achieve similar accuracy on faces with extreme variations in viewpoints, resolution, occlusion and image quality.  This is evident from the performance of the traditional algorithms on the publicly available IJB-A~\cite{klare2015pushing} dataset. Data quality imbalance in the training set is one of the reason for this performance gap. Existing face recognition training datasets contain large amount of high quality and frontal faces, whereas the unconstrained and difficult faces occur rarely. Most of the DCNN-based methods trained with softmax loss for classification tend to over-fit to the high quality data and fail to correctly classify faces acquired in difficult conditions.

Using softmax loss function for training face verification system has its own pros and cons. On the one hand, it can be easily implemented using inbuilt functions from the publicly available deep leaning toolboxes such as Caffe~\cite{jia2014caffe}, Torch~\cite{torch} and TensorFlow~\cite{tensorflow2015-whitepaper}. Unlike triplet loss~\cite{schroff2015facenet}, it does not have any restrictions on the input batch size and converges quickly. The learned features are discriminative enough for efficient face verification without any metric learning. 

On the other hand, the softmax loss is biased to the sample distribution. Unlike contrastive loss~\cite{sun2015deeply} and triplet loss~\cite{schroff2015facenet} which specifically attend to hard samples, the softmax loss maximizes the conditional probability of all the samples in a given mini-batch. Hence, it fits well to the high quality faces, ignoring the rare difficult faces from a training mini-batch. We observe that the $L_{2}$-norm of features learned using softmax loss is informative of the quality of the face~\cite{parde2016deep}. Features for good quality frontal faces have a high $L_{2}$-norm while blurry faces with extreme pose have low $L_{2}$-norm (see Figure~\ref{fig:verif_norms}(b)). Moreover, the softmax loss does not optimize the verification requirement of keeping positive pairs closer and negative pairs far from each other. Due to this reason, many methods either apply metric learning on top of softmax features~\cite{sankaranarayanan2016triplet, chen2016unconstrained, parkhi2015deep} or train an auxiliary loss~\cite{wen2016discriminative, sun2015deeply, wen2016latent}  along with the softmax loss to achieve better verification performance. 

%A simple experiment on IJB-A verification protocol suggests that template pairs with higher $L_{2}$-norm perform exceptionally better than template pairs with lower $L_{2}$-norm (Figure~\ref{fig:verif_norms}(a)).

% What is your solution
In this paper, we provide a symptomatic treatment to issues associated with the softmax loss. We propose an $L_{2}$-softmax loss that adds a constraint on the features during training such that their $L_{2}$-norm remain constant. In other words, we restrict the features to lie on a hypersphere of a fixed radius. The proposed $L_{2}$-softmax loss has a dual advantage. Firstly, it provides similar attention to both good and bad quality faces since all the features have the same $L_{2}$-norm now, which is essential for better performance in unconstrained settings. Secondly, it strengthens the verification signal by forcing the same subject features to be closer and different subject features to be far from each other in the normalized space. Thus, it maximizes the margin for the normalized $L_{2}$ distance or cosine similarity score between negative and positive pairs. Thus, it overcomes the main disadvantages of the regular softmax loss.

The $L_{2}$-softmax loss also retains the advantages of the regular softmax loss. Similar to the softmax loss, it is a one network, one loss system. It doesn't necessarily require any joint supervision as used by many recent methods~\cite{wen2016discriminative, parkhi2015deep, wen2016latent, sun2015deeply}. It can be easily implemented using inbuilt functions from Caffe~\cite{jia2014caffe}, Torch~\cite{torch} and TensorFlow~\cite{tensorflow2015-whitepaper}, and converges very fast. It introduces just a single scaling parameter to the network. Compared to the regular softmax loss, the $L_{2}$-softmax loss gains a significant boost in the performance. It achieves new state-of-the-art results on IJB-A dataset, and competing results on LFW and YouTube Face datasets. It surpasses the performance of several state-of-the-art systems, which use multiple networks or multiple loss functions or both. In summary, this paper contributes to the following aspects:
 
\begin{enumerate}
\item We propose a simple, novel and effective $L_{2}$-softmax loss for face verification that restricts the $L_{2}$-norm of the feature descriptor to a constant value $\alpha$.
\item We study the variations in the performance with respect to the scaling parameter $\alpha$ and provide suitable bounds on its value for achieving consistently high performance.
\item The proposed method yields a consistent and significant boost on all the three challenging face verification datasets namely LFW~\cite{huang2007labeled}, YouTube Face~\cite{liu2015targeting} and IJB-A~\cite{klare2015pushing}
\end{enumerate}

Moreover, the gains from $L_{2}$-softmax loss are complementary to metric learning (eg: TPE~\cite{sankaranarayanan2016triplet}, joint-Bayes~\cite{chen2016unconstrained}) or auxiliary loss functions (eg: center loss~\cite{wen2016discriminative}, contrastive loss~\cite{sun2015deeply}). We show that applying these techniques on top of the $L_{2}$-softmax loss can further improve the verification performance. Combining with TPE~\cite{sankaranarayanan2016triplet}, $L_{2}$-softmax loss achieves a record True Accept Rate (TAR) of 0.909 at False Accept Rate (FAR) of 0.0001 on the challenging IJB-A~\cite{klare2015pushing} dataset.

%% file: relatedWork.tex
%!TEX root = nsface_for_review.tex

\section{Related Work}

In recent years, there has been a significant improvement in the accuracy of face verification using deep learning methods~\cite{schroff2015facenet, taigman2014deepface, parkhi2015deep, sankaranarayanan2016triplet, sun2015deeply, wen2016discriminative}. Most of these methods have even surpassed human performance on the LFW~\cite{huang2007labeled} dataset. Although these methods use DCNNs, they differ by the type of loss function they use for training. For face verification, its essential for the positive subjects pair features to be closer and negative subjects pair features far apart. To solve this problem, researchers have adopted two major approaches. 

In the first approach, pairs of face images are input to the training algorithm to learn a feature embedding where positive pairs are closer and negative pairs are far apart. In this direction, Chopra et al.~\cite{chopra2005learning} proposed siamese networks with contrastive loss for training. Hu et al.~\cite{hu2014discriminative} designed a  discriminative deep metric with
a margin between positive and negative face pairs. FaceNet~\cite{schroff2015facenet} introduces triplet loss to learn the metric using hard triplet face samples. 

In the second approach, the face images along with their subject labels are used to learn discriminative identification features in a classification framework. Most of the recent methods~\cite{sun2015deeply, taigman2014deepface, parkhi2015deep, yang2016neural} train a DCNN with softmax loss to learn these features which are later used either to directly compute the similarity score for a pair of faces or to train a discriminative metric embedding~\cite{sankaranarayanan2016triplet, chen2016unconstrained}. Another strategy is to train the network for joint identification-verification task~\cite{sun2015deeply, wen2016latent, wen2016discriminative}. 
Xiong et al.~\cite{xiong2017good} proposed transferred deep feature fusion (TDFF) which infolves two-stage fusion of features trained with different networks and datasets. Template adaptation~\cite{crosswhite2016template} is applied to further boost the performance.

A recent approach~\cite{wen2016discriminative} introduced center loss to learn better discriminative face features. Our proposed method is different from the center loss in the following aspects. First, we use one loss function (i.e., $L_{2}$-softmax loss) whereas~\cite{wen2016discriminative} uses center loss jointly with the softmax loss during training. Second, center loss introduces $C \times D$ additional parameters during training where $C$ is the number of classes and $D$ is the feature dimension. On the other hand, the $L_{2}$-softmax loss introduces just a single parameter that defines the fixed $L_{2}$-norm of the features. Moreover, the center loss can also be used in conjunction with $L_{2}$-softmax loss, which performs better than center loss trained with regular softmax loss (see Section~\ref{sec:centerloss}).

Recently, a few algorithms have used feature normalization during training to improve performance. SphereFace~\cite{liu2017sphereface} proposes angular softmax (A-softmax) loss that enables DCNNs to learn angularly discriminative features. Another method called DeepVisage~\cite{hasnat2017deepvisage} uses a special case of batch normalization technique to normalize the feature descriptor before applying the softmax loss. Our proposed method is different as it applies an $L_{2}$-constraint on the feature descriptors enforcing them to lie on a hypersphere of a given radius.

%% file: motivation.tex
%!TEX root = nsface_for_review.tex

\section{Motivation}

\begin{figure*}[htp!]
 \centering
\includegraphics[width=6.0cm, height=4.5cm]{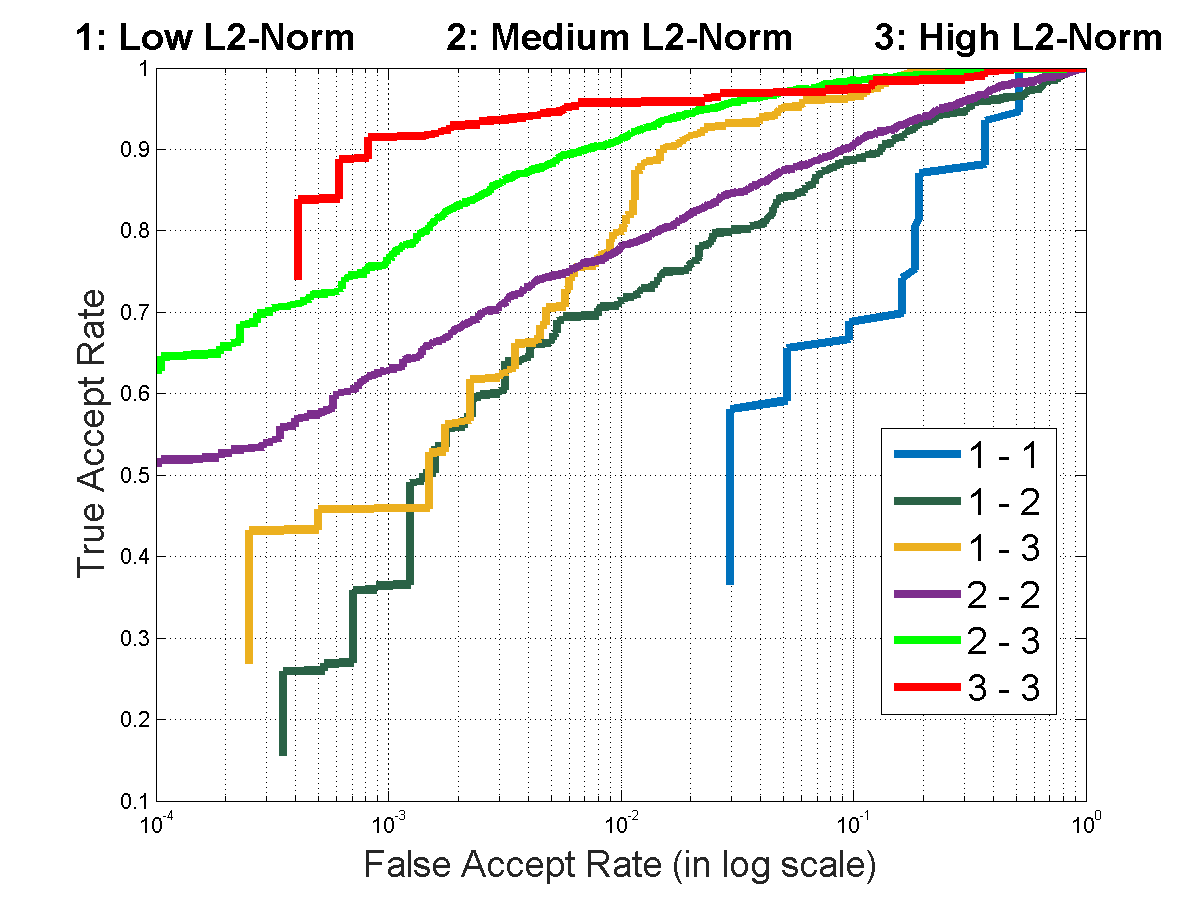}\hskip20pt\includegraphics[width=10.0cm, height=4.5cm]{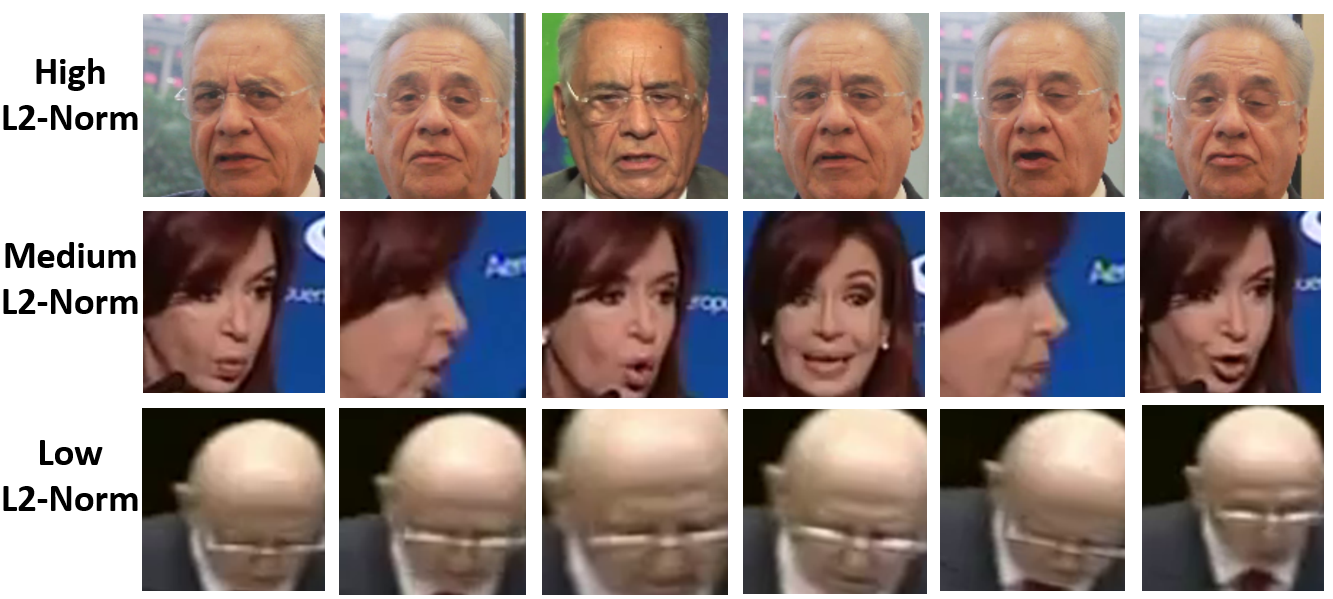}\\
(a)\hskip270pt(b)
\caption{(a) Face Verification Performance on IJB-A dataset. The templates are divided into $3$ sets based on their $L_{2}$-norm. `1' denotes the set with low $L_{2}$-norm while `3' represents high $L_{2}$-norm. The legend `x-y' denote the evaluation pairs where one template is from set `x' and another from set `y'. (b) Sample template images from IJB-A dataset with high, medium and low L2-norm}
\label{fig:verif_norms}
\end{figure*}

\begin{figure}[htp!]
      \centering
      \includegraphics[width=7.0cm, height=4.5cm]{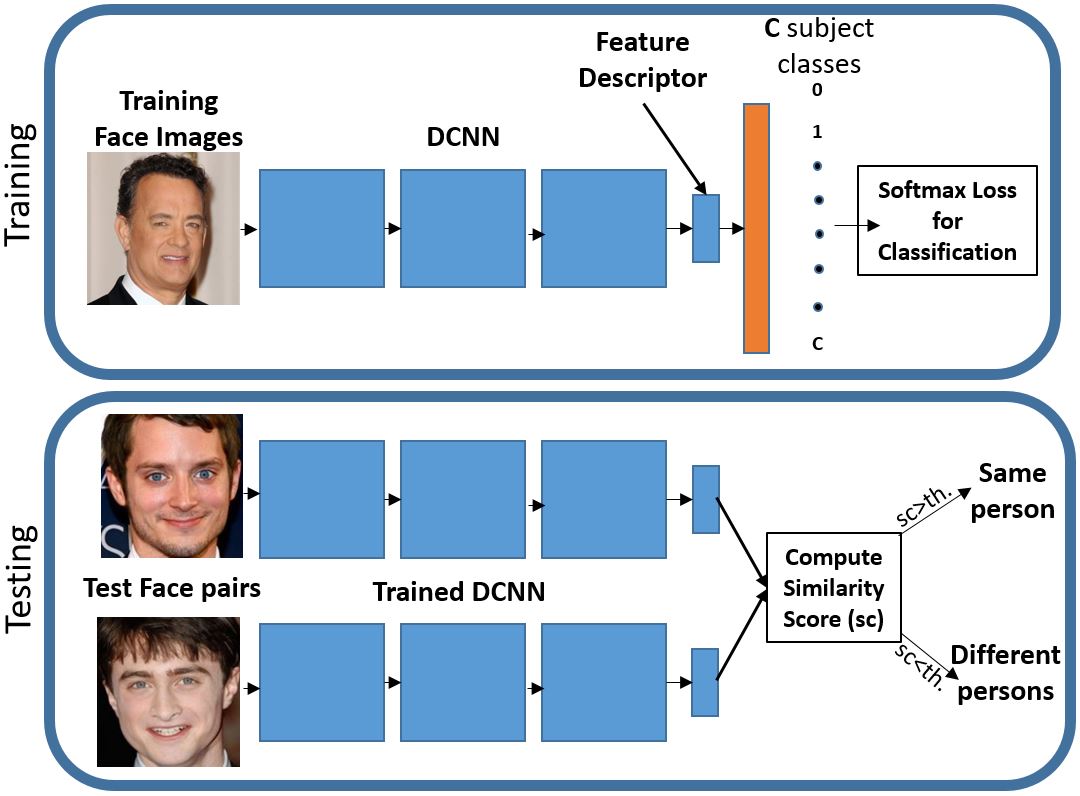}
      \caption{A general pipeline for training and testing a face verification system using DCNN.}
      \label{fig:pipeline}
\end{figure}

We first summarize the general pipeline for training a face verification system using DCNN as shown in Figure~\ref{fig:pipeline}. Given a training dataset with face images and corresponding identity labels, a DCNN is trained as a classification task where the network learns to classify a given face image to its correct identity label. A softmax loss function is used for training the network, given by Equation~\ref{eq:softmax_loss}

\begin{equation}
\label{eq:softmax_loss}
L_{S} = - \frac{1}{M} \sum_{i=1}^{M} \log \frac{e^{W_{y_{i}}^{T}f(\mathbf{x}_{i})+b_{y_{i}}}}{\sum_{j=1}^{C} e^{W_{j}^{T}f(\mathbf{x}_{i})+b_{j}}},
\end{equation}  

where $M$ is the training batch size, $\mathbf{x}_{i}$ is the $i^{th}$ input face image in the batch, $f(\mathbf{x}_{i})$ is the corresponding output of the penultimate layer of the DCNN, $y_{i}$ is the corresponding class label, and $W$ and $b$ are the weights and bias for the last layer of the network which acts as a classifier.

At test time, feature descriptors $f(\mathbf{x}_{g})$ and $f(\mathbf{x}_{p})$ are extracted for the pair of test face images $\mathbf{x}_{g}$ and $\mathbf{x}_{p}$ respectively using the trained DCNN, and normalized to unit length. Then, a similarity score is computed on the feature vectors which provides a measure of distance or how close the features lie in the embedded space. If the similarity score is greater than a set threshold, the face pairs are decided to be of the same person. Usually, the similarity score is computed as the $L_{2}$-distance between the normalized features~\cite{schroff2015facenet, parkhi2015deep} or by using cosine similarity~\cite{wen2016discriminative, chen2016unconstrained, ranjan2016all, sankaranarayanan2016triplet} $s$, as given by Equation~\ref{eq:cosine_similarity}. Both these similarity measures are equivalent and produce same results.

\begin{equation}
\label{eq:cosine_similarity}
s = \frac{f(\mathbf{x}_{g})^{T} f(\mathbf{x}_{p})}{\|f(\mathbf{x}_{g})\|_2 \|f(\mathbf{x}_{p})\|_2}
\end{equation} 

There are two major issues with this pipeline. First, the training and testing steps for face verification task are decoupled. Training with softmax loss doesn't necessarily ensure the positive pairs to be closer and the negative pairs to be far separated in the normalized or angular space. 

%Firstly, unlike the traditional image classification problem where a test sample needs to be classified into one of the pre-defined C classes, the training and testing steps for face verification task are decoupled.
%To incorporate this discriminative information during training, researches often use auxiliary loss functions such as Triplet Loss~\cite{schroff2015facenet,sankaranarayanan2016triplet}, Contrastive Loss, Center Loss~\cite{wen2016discriminative}, etc in addition to the softmax loss.

Secondly, the softmax classifier is weak in modeling difficult or extreme samples. In a typical training batch with data quality imbalance, the softmax loss gets minimized by increasing the $L_{2}$-norm of the features for easy samples, and ignoring the hard samples. The network thus learns to respond to the quality of the face by the $L_{2}$-norm of its feature descriptor. To validate this theory, we perform a simple experiment on the IJB-A~\cite{klare2015pushing} dataset where we divide the templates (groups of images/frames of same subject) into three different sets based on the $L_{2}$-norm of their feature descriptors. The features were computed using Face-Resnet~\cite{wen2016discriminative} trained with regular softmax loss. Templates with descriptors' $L_{2}$-norm \textless $90$ are assigned to set$1$. The templates with $L_{2}$-norm \textgreater $90$ but \textless $150$ are assigned to set$2$, while templates with $L_{2}$-norm \textgreater $150$ are assigned to set$3$. In total they form six sets of evaluation pairs. Figure~\ref{fig:verif_norms}(a) shows the performance of the these six different sets for the IJB-A face verification protocol. It can be clearly seen that pairs having low $L_{2}$-norm for both the templates perform very poor, while the pairs with high $L_{2}$-norm perform the best. The difference in performance between each set is quite significant. Figure~\ref{fig:verif_norms}(b) shows some sample templates from set$1$, set$2$ and set$3$ which confirms that the $L_{2}$-norm of the feature descriptor  is informative of its quality.

To solve these issues, we enforce the $L_{2}$-norm of the features to be fixed for every face image. Specifically, we add an $L_{2}$-constraint to the feature descriptor such that it lies on a hypersphere of a fixed radius. This approach has two advantages. Firstly, on a hypersphere, minimizing the softmax loss is equivalent to maximizing the cosine similarity for the positive pairs and minimizing it for the negative pairs, which strengthens the verification signal of the features. Secondly, the softmax loss is able to model the extreme and difficult faces better, since all the face features have same $L_{2}$-norm.

%% file: method.tex
%!TEX root = nsface_for_review.tex

\section{Proposed Method}
\label{sec:method}
The proposed $L_{2}$-softmax loss is given by Equation~\ref{eq:l2_softmax_loss}

\begin{equation}
\label{eq:l2_softmax_loss}
\begin{aligned}
{\text{minimize}} && - \frac{1}{M} \sum_{i=1}^{M} \log \frac{e^{W_{y_{i}}^{T}f(\mathbf{x}_{i})+b_{y_{i}}}}{\sum_{j=1}^{C} e^{W_{j}^{T}f(\mathbf{x}_{i})+b_{j}}} \\
\text{subject to} && \|f(\mathbf{x}_{i})\|_2 = \alpha, ~~\forall i = 1,2,... M,\\
\end{aligned}
\end{equation}  

where $\mathbf{x}_{i}$ is the input image in a mini-batch of size $M$, $y_{i}$ is the corresponding class label, $f(\mathbf{x}_{i})$ is the feature descriptor obtained from the penultimate layer of DCNN, $C$ is the number of subject classes, and $W$ and $b$ are the weights and bias for the last layer of the network which acts as a classifier. This equation adds an additional $L_{2}$-constraint to the regular softmax loss defined in Equation~\ref{eq:softmax_loss}. We show the effectiveness of this constraint using MNIST~\cite{lecun1998mnist} data.

%\begin{equation}
%\begin{aligned}
%\underset{\mathbf{x} \in \mathbb{R}^{n}}{\text{minimize}}
%&&\mathbf{x}^{T}\mathbf{A}\mathbf{x}\\
%\text{subject to} &&\|\mathbf{x}\|_\infty ~\leq~ 1;\\
%&&\|\mathbf{x}\|_1 ~\geq~ n
%\label{EquivL1BQP}
%\end{aligned}
%\end{equation}

%We propose a simple yet effective $L_{2}$-softmax loss which adds an $L_{2}$-norm constraint on the deep feature descriptors such that they lie on a hypersphere of a given radius. 

\subsection{MNIST Example}
\label{sec:toy_example}

\begin{figure}[htp!]
      \centering
\includegraphics[width=3.8cm, height=3.8cm]{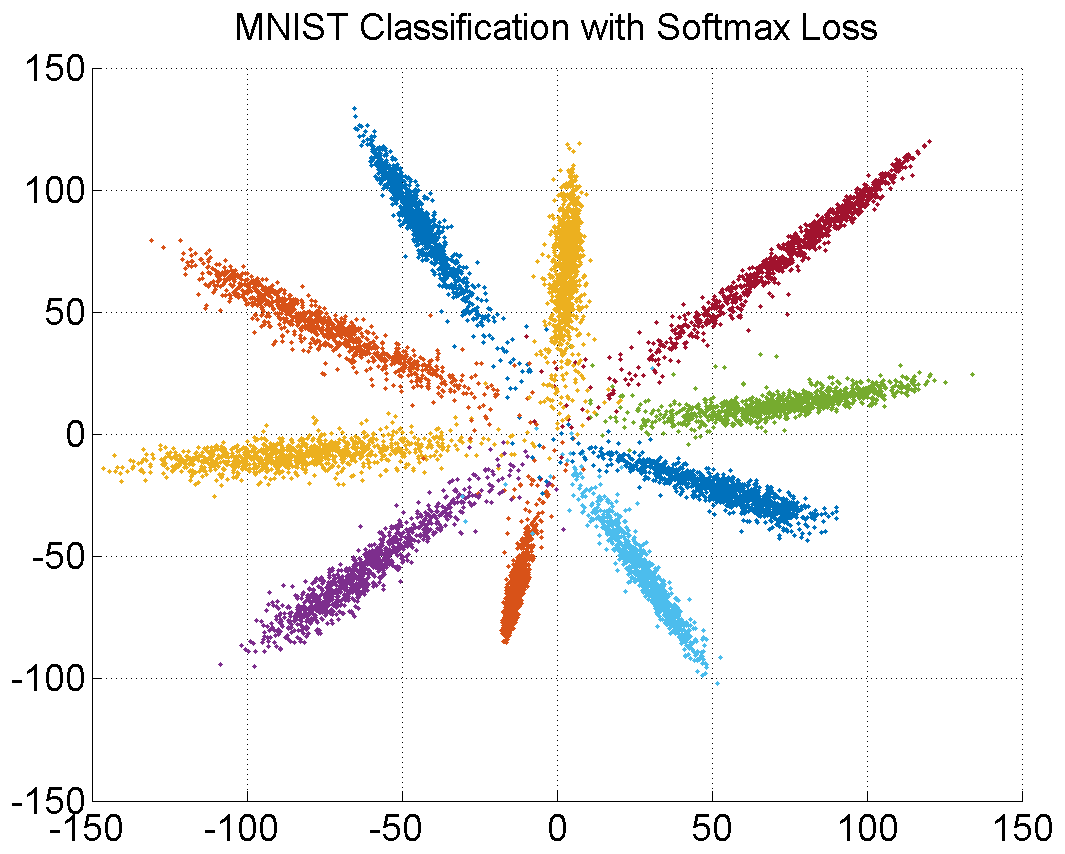}\hskip20pt\includegraphics[width=3.8cm, height=3.8cm]{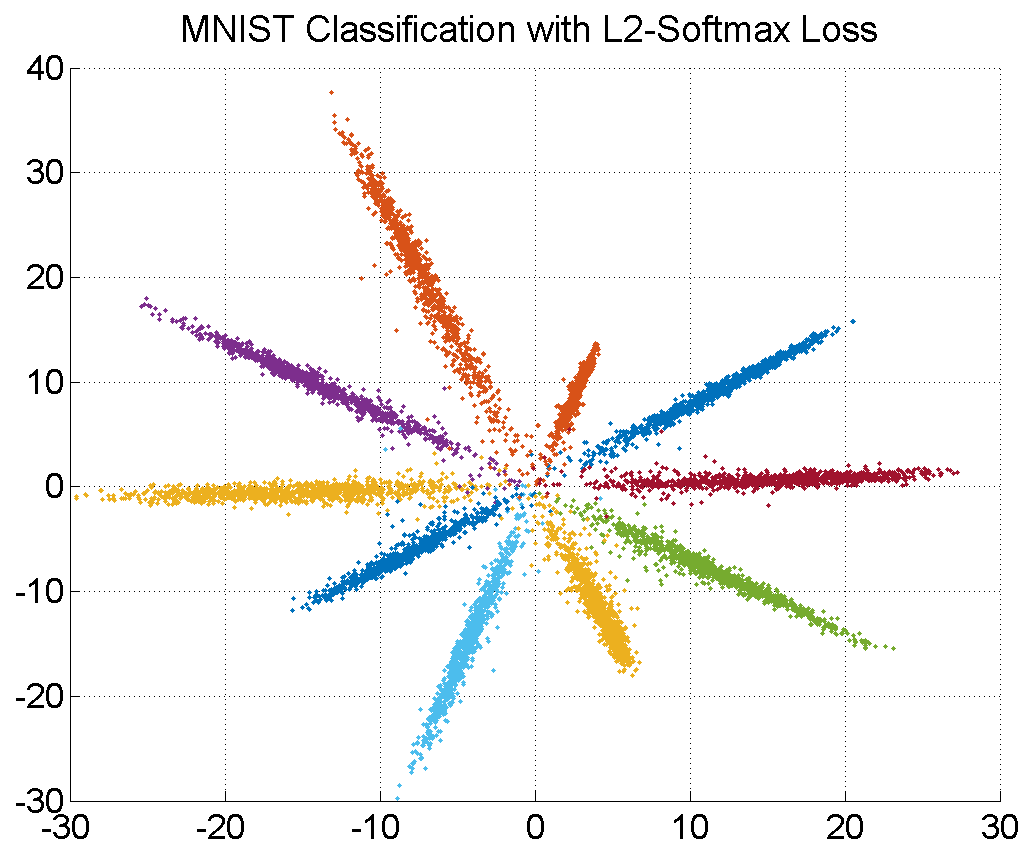}\\
(a)\hskip120pt(b)
\caption{Vizualization of $2$-dimensional features for MNIST digit classification test set using (a) Softmax Loss. (b) L2-Softmax Loss}
      \label{fig:mnist}
\end{figure}

We study the effect of $L_{2}$-softmax loss on the MNIST dataset~\cite{lecun1998mnist}. We use a deeper and wider version of LeNet mentioned in~\cite{wen2016discriminative}, where the last hidden layer output is restricted to $2$-dimensions for easy vizualization. For the first setup, we train the network end-to-end using the regular softmax loss for digits classifcation with number of classes = $10$. For the second setup, we add an $L_{2}$-normalize layer and scale layer to the $2$-dimensional features which enforces the $L_{2}$-constraint described in Equation~\ref{eq:l2_softmax_loss} (seen Section~\ref{sec:implementation} for details). Figure~\ref{fig:mnist} depicts the $2$-D features for different classes for MNIST test set containing $10,000$ digit images. Each of the lobes shown in the figure represents $2$-D features of unique digits classes. The features for the second setup were obtained before the $L_{2}$-normalization layer.

\begin{table}[htp!]
\centering
\caption{Accuracy on MNIST test set in (\%)}
\label{tbl:mnist}
%\begin{minipage}{.25\linewidth}
%%\centering
%\tabcolsep=0.15cm
\begin{tabular}{|c|c|c|}
\hline
~ & Softmax Loss & L2-Softmax Loss\\
\hline
Accuracy&98.88&99.05\\
\hline
\end{tabular}
\end{table}

We find two clear differences between the features learned using the two setups discussed above. First, the intra-class angular variance is large when using the regular softmax loss, which can be estimated by the average width of the lobes for each class. On the other hand, the features obtained with $L_{2}$-softmax loss have lower intra-class angular variability, and are represented by thinner lobes. Second, the magnitudes of the features are much higher with the softmax loss (ranging upto $150$), since larger feature norms result in a higher probability for a correctly classified class. In contrast, the feature norm has minimal effect on the $L_{2}$-softmax loss since every feature is normalized to a circle of fixed radius before computing the loss. Hence, the network focuses on bringing the features from the same class closer to each other and separating the features from different classes in the normalized or angular space. Table~\ref{tbl:mnist} lists the accuracy obtained with the two setups on MNIST test set. $L_{2}$-softmax loss achieves a higher performance, reducing the error by more than $15\%$. Note that these accuracy numbers are lower compared to a typical DCNN since we are using only $2$-dimensional features for classification.

\subsection{Implementation Details}
\label{sec:implementation}
\begin{figure}[htp!]
      \centering
      \includegraphics[width=0.5\textwidth]{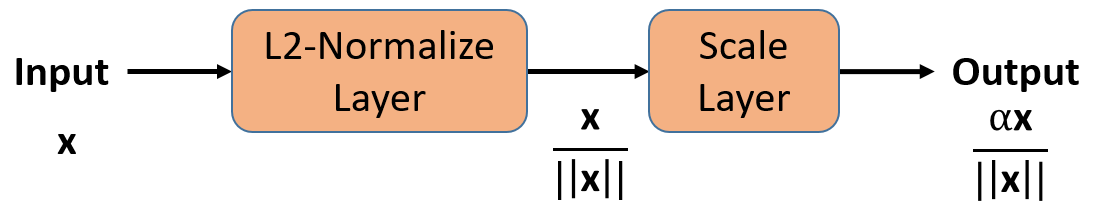}
      \caption{We add an $L_{2}$-normalize layer and a scale layer to constrain the feature descriptor to lie on a hypersphere of radius $\alpha$.}
      \label{fig:model}
\end{figure}

Here, we provide the details of implementing the $L_{2}$-constraint described in Equation~\ref{eq:l2_softmax_loss} in the framework of DCNNs. The constraint is enforced by adding an $L_{2}$-normalize layer followed by a scale layer as shown in Figure~\ref{fig:model}.

This module is added just after the penultimate layer of DCNN which acts as a feature descriptor. The $L_{2}$-normalize layer normalizes the input feature $\mathbf{x}$ to a unit vector given by Equation~\ref{eq:l2_normalize}. The scale layer scales the input unit vector to a fixed radius given by the parameter $\alpha$ (Equation~\ref{eq:scale}). In total, we just introduce one scalar parameter ($\alpha$) which can be trained along with the other parameters of the network.

\begin{equation}
\label{eq:l2_normalize}
\mathbf{y} = \frac{\mathbf{x}}{\|\mathbf{x}\|_2}
\end{equation}  

\begin{equation}
\label{eq:scale}
\mathbf{z} = \alpha \cdot \mathbf{y}
\end{equation}  

%The module is fully differentiable and can backpropagate the gradients to the lower DCNN layers during training using the chain rule. In order to compute the gradient through $L_{2}$-normalize layer for backpropagation, we rewrite Equation~\ref{eq:l2_normalize} for a single output unit $y_{i}$ in terms of the input vector $\mathbf{x}$ with dimension $D$ and proceed accordingly as follows:
The module is fully differentiable and can be used in the end-to-end training of the network. At test time, the proposed module is redundant, since the features are eventually normalized to unit length while computing the cosine similarity. At training time, we backpropagate the gradients through the L2-normalize and the scale layer,  as well as compute the gradients with respect to the scaling parameter $\alpha$ using the chain rule as given below.

\begin{equation}
\begin{aligned}
\frac{\partial l}{\partial y_{i}} &= \frac{\partial l}{\partial z_{i}} \cdot \alpha \\
\frac{\partial l}{\partial \alpha} &= \sum_{j=1}^{D} \frac{\partial l}{\partial z_{j}} \cdot y_{j} \\
\frac{\partial l}{\partial x_{i}} &= \sum_{j=1}^{D} \frac{\partial l}{\partial y_{j}} \cdot \frac{\partial y_{j}}{\partial x_{i}} \\
\frac{\partial y_{i}}{\partial x_{i}} &= \frac{\|\mathbf{x}\|_2^{2} - x_{i}^{2}}{\|\mathbf{x}\|_2^{3}} \\
\frac{\partial y_{j}}{\partial x_{i}} &= \frac{- x_{i} \cdot x_{j}}{\|\mathbf{x}\|_2^{3}}
\end{aligned}
\end{equation}

%\begin{equation}
%y_{i} = \frac{x_{i}}{\sqrt{\sum_{j=1}^{D} x_{j}^{2}}}
%\end{equation}
%
%\begin{equation}
%\begin{aligned}
%\frac{dy_{i}}{dx_{i}} &= \frac{\sqrt{\sum_{j=1}^{D} x_{j}^{2}} -  \frac{x_{i}^{2}}{\sqrt{\sum_{j=1}^{D} x_{j}^{2}}}}{\sum_{j=1}^{D} x_{j}^{2}}\\
%&= \frac{\|\mathbf{x}\|_2^{2} - x_{i}^{2}}{\|\mathbf{x}\|_2^{3}}\\
%\frac{dy_{j}}{dx_{i}} &= \frac{- x_{i} \cdot x_{j}}{\|\mathbf{x}\|_2^{3}}
%\end{aligned}
%\end{equation}
%
%The gradient through scale layer simply gets multiplied with the scalar parameter $\alpha$ during back-propagation. The gradient for the scale parameter $\alpha$ is given by Equation~\ref{eq:alpha_update}.
%
%\begin{equation}
%\begin{aligned}
%\Delta \alpha &= \sum_{j=1}^{D} \frac{dz_{j}}{d \alpha}\\
% &= \sum_{j=1}^{D} y_{j}
%\end{aligned}
%\label{eq:alpha_update}
%\end{equation}

\subsection{Bounds on Parameter $\alpha$}
\label{sec:theory}

The scaling parameter $\alpha$ plays a crucial role in deciding the performance of $L_{2}$-softmax loss. There are two ways to enforce the $L_{2}$-constraint: 1) by keeping $\alpha$ fixed throughout the training, and 2) by letting the network to learn the parameter $\alpha$. The second way is elegant and always improves over the regular softmax loss. But, the $\alpha$ parameter learned by the network is high which results in a relaxed $L_{2}$-constraint. The softmax classifier aimed at increasing the feature norm for minimizing the overall loss, increases the $\alpha$ parameter instead, allowing it more freedom to fit to the easy samples. Hence, $\alpha$ learned by the network forms an upper bound for the parameter. A better performance is obtained by fixing $\alpha$ to a lower constant value.

On the other hand, with a very low value of $\alpha$, the training doesn't converge. For instance, $\alpha = 1$ performs very poorly on the LFW~\cite{huang2007labeled} dataset, achieving an accuracy of $86.37\%$ (see Figure~\ref{fig:resnet_small}). The reason being that a hypersphere with small radius ($\alpha$) has limited surface area for embedding features from the same class together and those from different classes far from each other.

\begin{figure}[htp!]
      \centering
\includegraphics[width=3.0cm, height=3.0cm]{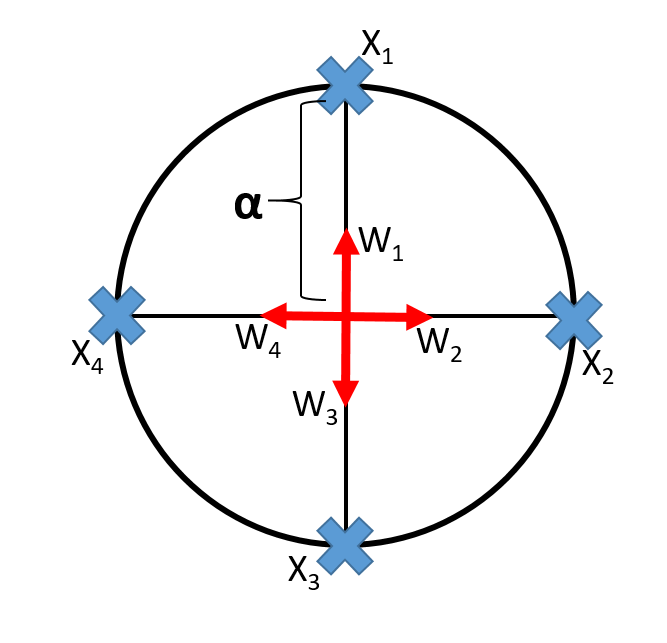}\hskip20pt\includegraphics[width=4.2cm, height=3.5cm]{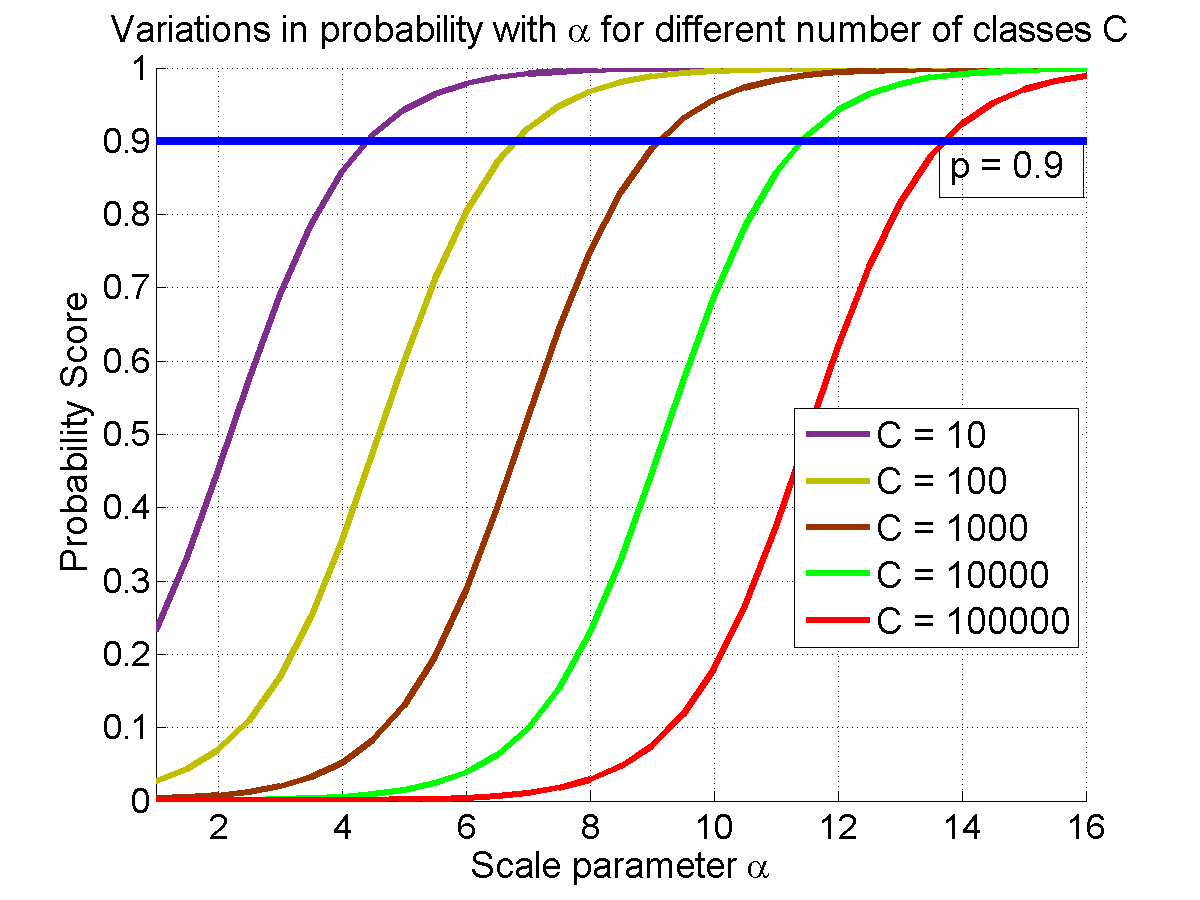}\\
(a)\hskip120pt(b)
\caption{(a) $2$-D vizualization of the assumed distribution of features (b) Variation in Softmax probability with respect to $\alpha$ for different number of classes $C$}
      \label{fig:lower_bound}
\end{figure}

Here, we formulate a theoretical lower bound on $\alpha$. Assuming the number of classes $C$ to be lower than twice the feature dimension $D$, we can distribute the classes on a hypersphere of dimension $D$ such that any two class centers are at least $90^{\circ}$ apart. Figure~\ref{fig:lower_bound}(a) represents this case for $C=4$ class centers distributed on a circle of radius $\alpha$. We assume the classifier weights ($W_{i}$) to be a unit vector pointing in the direction of their respective class centers. We ignore the bias term. The average softmax probability $p$ for correctly classifying a feature is given by Equation~\ref{eq:2d_alpha}

\begin{equation}
\label{eq:2d_alpha}
\begin{aligned}
p &= \frac{e^{W_{i}^T X_{i}}}{\sum_{j=1}^{4} e^{W_{j}^T X_{i}}}\\
&= \frac{e^{\alpha}}{e^{\alpha} + 2 + e^{-\alpha}}
\end{aligned}
\end{equation}  

Ignoring the term $ e^{-\alpha}$ and generalizing it for $C$ classes, the average probability becomes:

\begin{equation}
\label{eq:2d_alpha1}
\begin{aligned}
p &= \frac{e^{\alpha}}{e^{\alpha} + C - 2}
\end{aligned}
\end{equation}

Figure~\ref{fig:lower_bound}(b) plots the probability score as a function of the parameter $\alpha$ for various number of classes $C$. We can infer that to achieve a given classification probability (say $p = 0.9$), we need to have a higher $\alpha$ for larger $C$. Given the number of classes $C$ for a dataset, we can obtain the lower bound on $\alpha$ to achieve a probability score of $p$ by using Equation~\ref{eq:lower_bound}.

\begin{equation}
\label{eq:lower_bound}
\alpha_{low} = \log \frac{p (C-2)}{1-p}
\end{equation}

%% file: results.tex
%!TEX root = nsface_for_review.tex

\section{Results}
\label{sec:results}

\begin{figure*}[htp!]
 \centering
\includegraphics[width=18.0cm, height=3.0cm]{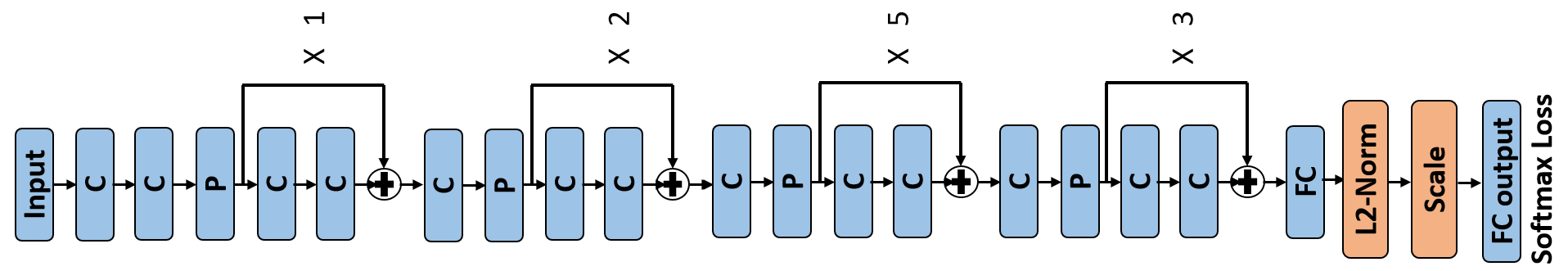}
\caption{The Face-Resnet architecture~\cite{wen2016discriminative} used for the experiments. \textbf{C} denotes Convolution Layer followed by PReLU~\cite{he2015delving} while \textbf{P} denotes Max Pooling Layer. Each pooling layer is followed by a set of residual connections, the count for which is denoted alongside. After the fully-connected layer (\textbf{FC}), we add an $L_{2}$-Normalize layer and Scale Layer which is followed by the softmax loss.}
\label{fig:architecture}
\end{figure*}

We use the publicly available Face-Resnet~\cite{wen2016discriminative} DCNN for our experiments. Figure~\ref{fig:architecture} shows the architecture of the network. It contains $27$ convolutional layers and $2$ fully-connected layers. The dimension of the feature descriptor is $512$. It utilizes the widely used residual skip-connections~\cite{he2016deep}. We add an $L_{2}$-normalize layer and a scale layer after the fully-connected layer to enforce the $L_{2}$-constraint on the descriptor. All our experiments are carried out in Caffe~\cite{jia2014caffe}.

\subsection{Baseline experiments}
In this subsection, we experimentally validate the usefulness of the $L_{2}$-softmax loss for face verification. We form two subsets of training dataset from the MS-Celeb-1M~\cite{guo2016ms} dataset: 1) MS-small containing $0.5$ million face images with the number of subjects being $13403$, and 2) MS-large containing $3.7$ million images of $58207$ subjects. The dataset was cleaned using the clustering algorithm mentioned in~\cite{lin2017proximity}. We train the Face-Resnet network with softmax loss as well as $L_{2}$-softmax loss for various $\alpha$. While training with MS-small, we start with a base learning rate of $0.1$ and decrease it by ${1/10}^{th}$ after $16K$ and $24K$ iterations, upto a maximum of $28K$ iterations. For training on MS-large, we use the same learning rate but decrease it after $50K$ and $80K$ iterations upto a maximum of $100K$ iterations. A training batch size of $256$ was used. Both softmax and $L_{2}$-softmax loss functions consume the same amount of training time which is around $9$ hours for MS-small and $32$ hours for MS-large training set respectively, on two TITAN X GPUs.  We set the learning multiplier and decay multiplier for the scale layer to $1$ for trainable $\alpha$, and $0$ for fixed $\alpha$ during the network training. We evaluate our baselines on the widely used LFW dataset~\cite{huang2007labeled} for the unrestricted setting, and the challenging IJB-A dataset~\cite{klare2015pushing} on the 1:1 face verification protocol. The faces were cropped and aligned to the size of $128 \times 128$ in both training and testing phases by implementing the face detection and alignment algorithm mentioned in~\cite{ranjan2016all} . 
 
\subsubsection{Experiment with small training set}
\begin{figure}[htp!]
      \centering
      \includegraphics[width=0.5\textwidth]{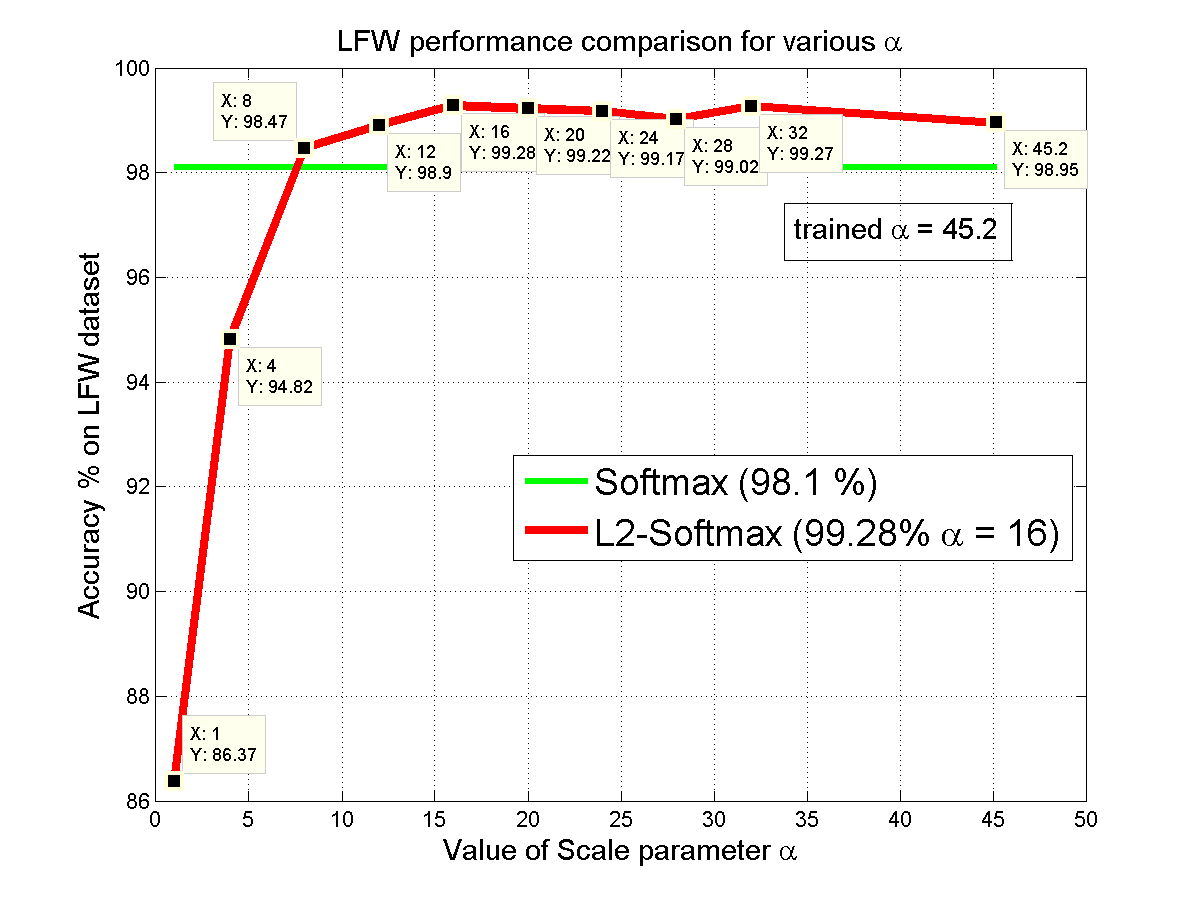}
      \caption{The red curve shows the variations in LFW accuracy with the parameter $\alpha$ for $L_{2}$-softmax loss. The green line is the accuracy using the usual softmax loss.}
      \label{fig:resnet_small}
\end{figure}

Here, we compare the network trained on MS-small dataset using our proposed $L_{2}$-softmax loss, against the one trained with regular softmax loss. Figure~\ref{fig:resnet_small} shows that the regular softmax loss attains an accuracy of $98.1\%$ whereas the proposed $L_{2}$-softmax loss achieves the best accuracy of $99.28\%$, thereby reducing the error by more than $62\%$. It also shows the variations in performance with the scale parameter $\alpha$. The performance is poor when $\alpha$ is below a certain threshold and stable with $\alpha$ higher than the threshold. This behavior is consistent with the theoretical analysis presented in Section~\ref{sec:theory}. From the figure, the performance of $L_{2}$-Softmax is better for $\alpha$ \textgreater $12$ which is close to its lower bound computed using equation~\ref{eq:lower_bound} for $C = 13403$ with a probability score of $0.9$. 

A similar trend is observed for 1:1 verification protocol on IJB-A~\cite{klare2015pushing} as shown in Table~\ref{tbl:resnet_small}, where the numbers denote True Accept Rate (TAR) at False Accept Rates (FAR) of $0.0001$, $0.001$, $0.01$ and $0.1$. Our proposed approach improves the TAR@FAR=0.0001 by $19\%$ compared to the baseline softmax loss. The performance is consistent with $\alpha$ ranging between $16$ to $32$. Another point to note is that by allowing the network to learn the scale parameter $\alpha$ by itself results in a slight decrease in performance, which shows that having a tighter constraint is a better choice.

\begin{table}[htp!]
\centering
\caption{TAR on IJB-A 1:1 Verification Protocol @FAR}
\label{tbl:resnet_small}
%\begin{minipage}{.25\linewidth}
%%\centering
\tabcolsep=0.15cm
\begin{tabular}{|c|c|c|c|c|}
\hline
~ & 0.0001 & 0.001 & 0.01 & 0.1\\
\hline
softmax & 0.553 & 0.730  & 0.881 & 0.957\\
\hline
\hline
$L_{2}$-softmax ($\alpha$=8) & 0.257 & 0.433  & 0.746 & 0.953\\
\hline
$L_{2}$-softmax ($\alpha$=12) & 0.620 & 0.721  & 0.875 & 0.970\\
\hline
$L_{2}$-softmax ($\alpha$=16) & 0.734 & 0.834  & \textbf{0.924} & 0.974\\
\hline
$L_{2}$-softmax ($\alpha$=20) & 0.740 & 0.820  & 0.922 & 0.973\\
\hline
$L_{2}$-softmax ($\alpha$=24) & \textbf{0.744} & 0.831  & 0.912 & 0.974\\
\hline
$L_{2}$-softmax ($\alpha$=28) & 0.740 & \textbf{0.834}  & 0.922 & \textbf{0.975}\\
\hline
$L_{2}$-softmax ($\alpha$=32) & 0.727 & 0.831  & 0.921 & 0.972\\
\hline
$L_{2}$-softmax ($\alpha$ trained) & 0.698 & 0.817  & 0.914 & 0.971\\
\hline
\end{tabular}
\end{table}

\subsubsection{Experiment with large training set}
\begin{figure}[htp!]
      \centering
      \includegraphics[width=0.5\textwidth]{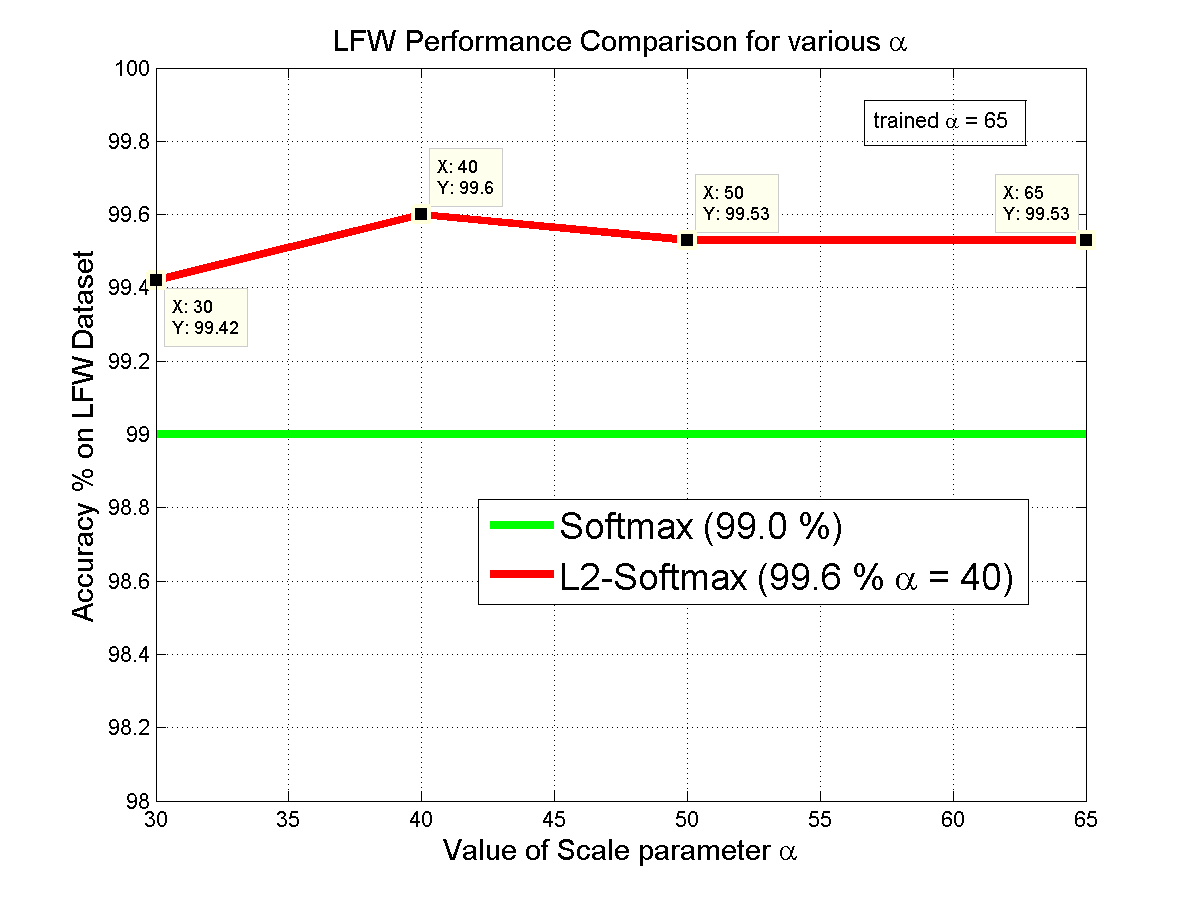}
      \caption{The red curve shows the variations in LFW accuracy with the parameter $\alpha$ for $L_{2}$-softmax loss. The green line is the accuracy using the softmax loss.}
      \label{fig:resnet_large}
\end{figure}

We train the network on the MS-large dataset for this experiment. Figure~\ref{fig:resnet_large} shows the performance on the LFW dataset. Similar to the small training set, the $L_{2}$-softmax loss significantly improves over the baseline, reducing the error by $60\%$ and achieving an accuracy of $99.6\%$. Similarly, it improves the TAR@FAR=0.0001 on IJB-A by more than $10\%$ (Table~\ref{tbl:resnet_large}). The performance of $L_{2}$-softmax is consistent with $\alpha$ in the range $40$ and beyond. Unlike, the small set training, the self-trained $\alpha$ performs equally good compared to fixed $\alpha$ of $40$ and $50$. The theoretical lower bound on $\alpha$ is not of much use in this case since improved performance is achieved for $\alpha$ \textgreater $30$. We can deduce that as the number of subjects increases, the lower bound on $\alpha$ is less reliable, and the self-trained $\alpha$ is more reliable with performance. This experiment clearly suggests that the proposed $L_{2}$-softmax loss is consistent across the training and testing datasets.

\begin{table}[htp!]
\centering
\caption{TAR on IJB-A 1:1 Verification Protocol @FAR}
\label{tbl:resnet_large}
%\begin{minipage}{.25\linewidth}
%%\centering
\tabcolsep=0.15cm
\begin{tabular}{|c|c|c|c|c|}
\hline
~ & 0.0001 & 0.001 & 0.01 & 0.1\\
\hline
softmax & 0.730 & 0.851  & 0.926 & 0.972\\
\hline
\hline
$L_{2}$-softmax ($\alpha$=30) & 0.775 & 0.871  & 0.938 & 0.978\\
\hline
$L_{2}$-softmax ($\alpha$=40) & 0.827 & 0.900  & 0.951 & \textbf{0.982}\\
\hline
$L_{2}$-softmax ($\alpha$=50) & \textbf{0.832} & \textbf{0.906}  & \textbf{0.952} & 0.981\\
\hline
$L_{2}$-softmax ($\alpha$ trained) & \textbf{0.832} & 0.903  & 0.950 & 0.980\\
\hline
\end{tabular}
\vspace*{-4mm}
\end{table}

\subsubsection{Experiment with a different DCNN}
\begin{figure}[htp!]
      \centering
      \includegraphics[width=0.5\textwidth]{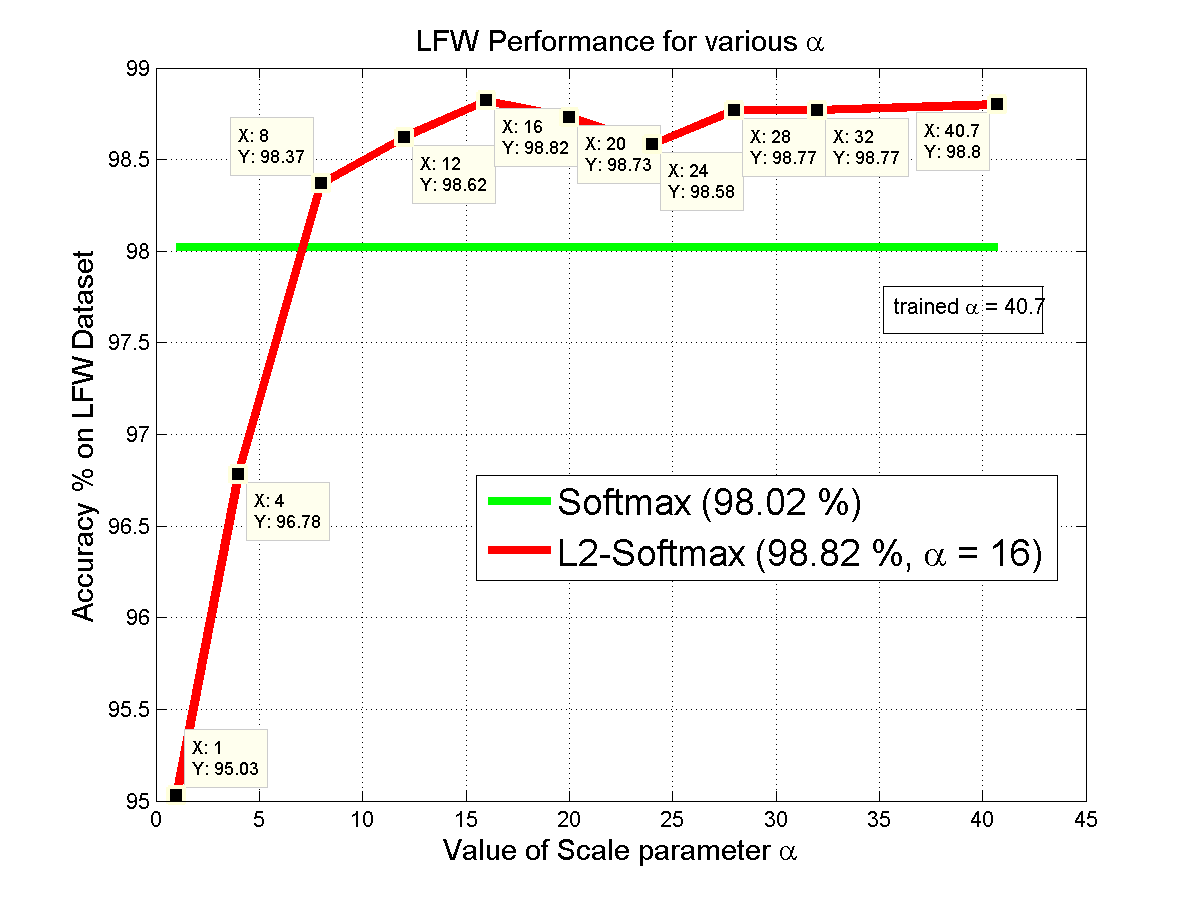}
      \caption{The red curve shows the variations in LFW accuracy with the parameter $\alpha$ for $L_{2}$-softmax loss. The green line is the accuracy using the softmax loss.}
      \label{fig:uf_small}
\end{figure}

To check the consistency of our proposed $L_{2}$-softmax loss, we apply it on the All-In-One Face~\cite{ranjan2016all} instead of the Face-Resnet. We use the recognition branch of the All-In-One Face to fine-tune on the MS-small training set. The recognition branch of All-In-One Face consists of $7$ convolution layers followed by $3$ fully-connected layers and a softmax loss. We add an $L_{2}$-normalize and a scale layer after the $512$ dimension feature descriptor. Figure~\ref{fig:uf_small} shows the comparison of $L_{2}$-softmax loss and the base softmax loss on LFW dataset. Similar to the Face-Resnet, All-In-One Face with $L_{2}$-softmax loss improves over the base softmax performance, reducing the error by $40\%$, and achieving an accuracy of $98.82\%$. The improvement obtained by using All-In-One Face is smaller compared to the Face-Resnet. This shows that residual connections and depth of the network generate better feature embedding on a hypersphere. The performance variation with scaling parameter $\alpha$ is similar to that of Face-Resnet, indicating that the optimal scale parameter does not depend on the choice of the network. 

\subsubsection{Experiment with auxiliary loss}
\label{sec:centerloss}
Similar to softmax loss, the $L_{2}$-softmax loss can be coupled with auxiliary losses such as center loss, contrastive loss, triplet loss, etc. to further improve the performance. Here we study the performance variation of $L_{2}$-softmax loss when coupled with the center loss. We use the MS-small dataset for training the networks. Table~\ref{tbl:centerL2} lists the accuracy obtained on LFW dataset by different loss functions. The softmax loss performs the worst. The center loss improves the performance significantly when trained in conjunction with the softmax loss, and is comparable to the $L_{2}$-softmax loss. Training center loss with the $L_{2}$-softmax loss gives the best performance of $99.33\%$ accuracy. This shows that $L_{2}$-softmax loss is as versatile as the regular softmax loss and can be used efficiently with other auxiliary loss functions.

\begin{table}[htp!]
\centering
\caption{Accuracy on LFW (\%)}
\label{tbl:centerL2}
%\begin{minipage}{.25\linewidth}
%%\centering
%\tabcolsep=0.15cm
\begin{tabular}{|c|c|}
\hline
softmax loss & 98.10\\
\hline
center loss~\cite{wen2016discriminative} + softmax loss & 99.23\\
\hline
$L_{2}$-softmax loss & 99.28\\
\hline
center loss~\cite{wen2016discriminative} + $L_{2}$-softmax loss & \textbf{99.33}\\
\hline
\end{tabular}
\end{table}

\subsection{Comparison with recent methods}

We compare our algorithm with recently reported face verification methods on LFW~\cite{huang2007labeled}, YouTube Face~\cite{wolf2011face} and IJB-A~\cite{klare2015pushing} datasets. We crop and align the images for all these datasets by implementing the algorithm mentioned in~\cite{ranjan2016all}. We train the Face-Resnet (FR) with $L_{2}$-softmax as well as regular softmax loss using the MS-large training set. Additionally, we train ResNet-101(R101)~\cite{he2016deep} and ResNeXt-101(RX101)~\cite{xie2016aggregated} deep networks for face recognition using MS-large training set with $L_{2}$-softmax loss. Both R101 and RX101 models were initialized with parameters pre-trained on ImageNet~\cite{ILSVRC15} dataset. A fully-connected layer of dimension $512$ was added before the $L_{2}$-softmax classifier. The scaling parameter was kept fixed with a value of $\alpha=50$. Experimental results on different datasets show that $L_{2}$-softmax works efficiently with deeper models.

The LFW dataset~\cite{huang2007labeled} contains $13,233$ web-collected images from $5749$ different identities. We evaluate our model following the standard protocol of unrestricted with labeled outside data. We test on 6,000 face pairs and report the experiment results in Table~\ref{tbl:lfw}. Along with the accuracy values, we also compare with the number of images, networks and loss functions used by the methods for their overall training. The proposed method attains the state-of-the-art performance with the RX101 model, achieving an accuracy of $99.78\%$. Unlike other methods which use auxiliary loss functions such as center loss and contrastive loss along with the primary softmax loss, our method uses a single loss training paradigm which makes it easier and faster to train. 

YouTube Face (YTF)~\cite{wolf2011face} dataset contains $3425$ videos of $1595$ different people, with an average length of $181.3$ frames per video. It contains $10$ folds of $500$ video pairs. We follow the standard verification protocol and report the average accuracy on splits with cross-validation in  Table~\ref{tbl:lfw}. We achieve the accuracy of $96.08\%$ using $L_{2}$-softmax loss with RX101 network. Our method outperforms many recent algorithms and is only behind DeepVisage~\cite{hasnat2017deepvisage} which uses larger number of training samples, and VGG Face~\cite{parkhi2015deep} which further uses a discriminative metric learning on YTF.

\begin{table}[htp!]
\centering
\caption{Verification accuracy (in $\%$) of different methods on LFW and YTF datasets.}
\label{tbl:lfw}
%\begin{minipage}{.25\linewidth}
%%\centering
\tabcolsep=0.10cm
\begin{tabular}{|c|c|c|c|c|c|}
\hline
Method & Images & $\#$nets &One loss & LFW & YTF\\
\hline
\hline
Deep Face~\cite{taigman2014deepface} & 4M & 3&No & $97.35$ & $91.4$\\
\hline
DeepID-2+~\cite{sun2015deeply} & - & 25&No & $99.47$ & $93.2$\\
\hline
FaceNet~\cite{schroff2015facenet} & 200M & 1&Yes & $99.63$ & $95.12$\\
\hline
VGG Face~\cite{parkhi2015deep} & 2.6M & 1&No & $98.95$ & $\mathbf{97.3}$\\
\hline
Baidu~\cite{liu2015targeting} & 1.3M & 1&No & $99.13$ & - \\
\hline
Wen et al.~\cite{wen2016discriminative} & 0.7M & 1&No & $99.28$ & $94.9$\\
\hline
NAN~\cite{yang2016neural} & 3M & 1&No & $-$ & $95.72$\\
\hline
DeepVisage~\cite{hasnat2017deepvisage} & 4.48M & 1&Yes & $99.62$ & $\mathbf{96.24}$\\
\hline
SphereFace~\cite{liu2017sphereface} & 0.5M & 1&Yes & $99.42$ & $95.0$\\
\hline
\hline
softmax(FR) & 3.7M & 1&Yes & $99.0$ & $93.82$\\
\hline
$L_{2}$-S~(FR) & 3.7M & 1&Yes & $99.60$ & $95.54$\\
\hline
$L_{2}$-S~(R101) & 3.7M & 1&Yes & $99.67$ & $96.02$\\
\hline
$L_{2}$-S~(RX101) & 3.7M & 1&Yes & $\mathbf{99.78}$ & $\mathbf{96.08}$\\
\hline
\end{tabular}
\end{table}

%\begin{table}[htp!]
%\centering
%\caption{Verification performance of different methods on LFW dataset.}
%\label{tbl:lfw}
%%\begin{minipage}{.25\linewidth}
%%%\centering
%\tabcolsep=0.10cm
%\begin{tabular}{|c|c|c|c|c|c|}
%\hline
%Method & Images & Networks & Single loss & LFW$\%$ & YTF$\%$\\
%\hline
%\hline
%Deep Face~\cite{taigman2014deepface} & 4M & 3  & No & $97.35$ & $91.4$\\
%\hline
%DeepID-2+~\cite{sun2015deeply} & - & 1  & No & $98.70$ &\\
%\hline
%DeepID-2+~\cite{sun2015deeply} & - & 25  & No & $99.47$ &\\
%\hline
%FaceNet~\cite{schroff2015facenet} & 200M & 1  & Yes & $99.63$ &\\
%\hline
%Deep FR~\cite{parkhi2015deep} & 2.6M & 1  & No & $98.95$ &\\
%\hline
%Baidu~\cite{liu2015targeting} & 1.3M & 1  & No & $99.13$ &\\
%\hline
%Wen et al.~\cite{wen2016discriminative} & 0.7M & 1  & No & $99.28$ &\\
%\hline
%\hline
%Softmax & 3.7M & 1  & Yes & $99.0$ & \\
%\hline
%\textbf{L2-Softmax} & 3.7M & 1  & Yes & $\mathbf{99.60}$ &\\
%\hline
%\end{tabular}
%\end{table}

The IJB-A dataset~\cite{klare2015pushing} contains $500$ subjects with a total of $25,813$ images including $5,399$ still images and $20,414$ video frames. It contains faces with extreme viewpoints, resolution and illumination which makes it more challenging than the commonly used LFW dataset. Given a template containing multiple faces of the same individual, we generate a common vector representation by media pooling the individual face descriptors, as explained in~\cite{sankaranarayanan2016triplet}. Table~\ref{tbl:ijba} lists the performance of recent DCNN-based methods on IJB-A dataset. We achieve state-of-the-art result for both verification and the identification protocols. Since the $L_{2}$-softmax loss can be coupled with any other auxiliary loss, we use the Triplet Probabilistic Embedding (TPE)~\cite{sankaranarayanan2016triplet} to learn a $128$-dimensional embedding using the training splits of IJB-A. It further improves the performance and achieves a record TAR of 0.909 @ FAR = 0.0001. To the best of our knowledge, we are the first ones to surpass TAR of 0.9 @ FAR of 0.0001 on IJB-A. Our method performs significantly better than existing methods in most of the other metrics as well. The results on LFW~\cite{huang2007labeled}, YTF~\cite{wolf2011face} and IJB-A~\cite{klare2015pushing} datasets clearly suggests the effectiveness of the proposed $L_{2}$-softmax loss.

\begin{table*}
\caption{Face Identification and Verification Evaluation on IJB-A dataset}
\label{tbl:ijba}
\begin{center}
\tabcolsep=0.10cm
\scalebox{0.875}{
\begin{tabular}{ccccccccc}
\hline
 & \multicolumn{4}{c}{IJB-A Verification (TAR@FAR)} & \multicolumn{4}{c}{IJB-A Identification}\\
\hline
Method & 0.0001 & 0.001 & 0.01 & 0.1 & FPIR=0.01 & FPIR=0.1 & Rank=1 & Rank=10\\
\hline
GOTS~\cite{klare2015pushing} & - & 0.2(0.008) & 0.41(0.014) & 0.63(0.023) & 0.047(0.02) & 0.235(0.03) & 0.443(0.02) & -\\
B-CNN~\cite{chowdhury2016one} & - & - & - & - & 0.143(0.027) & 0.341(0.032) & 0.588(0.02) & - \\
LSFS~\cite{wang2015face} & - & 0.514(0.06) & 0.733(0.034) &  0.895(0.013) & 0.383(0.063) & 0.613(0.032) & 0.820(0.024) & - \\
VGG-Face~\cite{parkhi2015deep} & - & 0.604(0.06) & 0.805(0.03) & 0.937(0.01) & 0.46(0.07) & 0.67(0.03) & 0.913(0.01) & 0.981(0.005)\\
$DCNN_{manual}$+metric~\cite{chen2015end} & - & - & 0.787(0.043) & 0.947(0.011) & - & - & 0.852(0.018) & 0.954(0.007) \\
Pose-Aware Models~\cite{masi2016pose} & - & 0.652(0.037) & 0.826(0.018) & - & -& -& 0.840(0.012) & 0.946(0.007) \\
Chen~et~al.~\cite{chen2016unconstrained} &- & - & 0.838(0.042) & 0.967(0.009) & 0.577(0.094) & 0.790(0.033) & 0.903(0.012) & 0.977(0.007)\\
Deep Multi-Pose~\cite{abdalmageed2016face} & - & - & 0.876 & 0.954 & 0.52 & 0.75 & 0.846 & 0.947 \\
Masi
et al.~\cite{masi2016we} & - & 0.725 & 0.886 & - & - & - & 0.906 & 0.977\\
Triplet Embedding~\cite{sankaranarayanan2016triplet} & - & 0.813(0.02) & 0.90(0.01) & 0.964(0.005) & 0.753(0.03) & 0.863(0.014) & 0.932(0.01) & 0.977(0.005)\\
Template Adaptation~\cite{crosswhite2016template} & - & 0.836(0.027) & 0.939(0.013) & 0.979(0.004) & 0.774(0.049) & 0.882(0.016) & 0.928(0.01) & 0.986(0.003)\\
All-In-One Face~\cite{ranjan2016all} & - & 0.823(0.02) & 0.922(0.01) & 0.976(0.004) & 0.792(0.02) & 0.887(0.014) & 0.947(0.008) & 0.988(0.003)\\
NAN~\cite{yang2016neural}  & - & 0.881(0.011) & 0.941(0.008) & 0.979(0.004) & 0.817(0.041) & 0.917(0.009) & 0.958(0.005) & 0.986(0.003)\\
TDFF~\cite{xiong2017good}  & 0.875(0.013) & 0.919(0.006) & 0.961(0.007) & 0.988(0.003) & 0.878(0.035) & 0.941(0.010) & 0.964(0.006) & \textbf{0.992(0.002)}\\
TDFF~\cite{xiong2017good}+TPE~\cite{sankaranarayanan2016triplet}  & 0.877(0.018) & 0.921(0.005) & 0.961(0.007) & \textbf{0.989(0.003)} & 0.881(0.039) & 0.940(0.009) & 0.964(0.007) & \textbf{0.992(0.003)}\\
\hline
softmax~(FR) &0.730(0.076) &  0.851(0.021) & 0.926(0.01) & 0.972(0.004) & 0.788(0.048) & 0.892(0.015) & 0.953(0.008) & 0.984(0.004)\\
$L_{2}$-S~(FR) &0.832(0.027) &  0.906(0.016) & 0.952(0.007) & 0.981(0.003) & 0.852(0.042) & 0.930(0.01) & 0.963(0.007) & 0.986(0.002)\\
$L_{2}$-S~(FR)+TPE~\cite{sankaranarayanan2016triplet} &0.863(0.012) &  0.910(0.013) & 0.951(0.006) & 0.979(0.003) & 0.873(0.024) & 0.931(0.01) & 0.961(0.007) & 0.983(0.003)\\
$L_{2}$-S~(R101) &0.879(0.028) &  0.937(0.008) & 0.967(0.005) & 0.984(0.002) & 0.895(0.055) & 0.953(0.007) & 0.973(0.005) & 0.987(0.003)\\
$L_{2}$-S~(R101)+TPE~\cite{sankaranarayanan2016triplet} &0.898(0.019) &  0.942(0.006) & 0.969(0.004) & 0.983(0.003) & 0.910(0.045) & \textbf{0.956(0.007)} & 0.971(0.005) & 0.986(0.003)\\
$L_{2}$-S~(RX101) &0.883(0.032) &  0.938(0.008) & 0.968(0.004) & 0.987(0.002) & 0.903(0.046) & 0.955(0.007) & \textbf{0.975(0.005)} & 0.990(0.002)\\
$L_{2}$-S~(RX101)+TPE~\cite{sankaranarayanan2016triplet} &\textbf{0.909(0.007)} &  \textbf{0.943(0.005)} & \textbf{0.970(0.004)} & 0.984(0.002) & \textbf{0.915(0.041)} & \textbf{0.956(0.006)} & 0.973(0.005) & 0.988(0.003)\\
\hline
\end{tabular}
}
\end{center}
\vspace*{-5mm}
\end{table*}
%\vspace*{-2.0cm}

%% file: conclusion.tex
%!TEX root = nsface_for_review.tex

\section{Conclusions}

In this paper, we added a simple, yet effective, $L_{2}$-constraint to the regular softmax loss for training a face verification system. The constraint enforces the features to lie on a hypersphere of a fixed radius characterized by parameter $\alpha$. We also provided bounds on the value of $\alpha$ for achieving a consistent performance. Experiments on LFW, YTF and IJB-A datasets show that the proposed $L_{2}$-softmax loss provides a significant and consistent boost over the regular softmax loss and achieves the state-of-the-art result on IJB-A~\cite{klare2015pushing} dataset. In conclusion, $L_{2}$-softmax loss is a valuable replacement for the existing softmax loss, for the task of face verification. In the future, we would further explore the possibility of exploiting the geometric structure of the feature encoding using manifold-based metric learning.